\documentclass[letterpaper]{article} 
\usepackage{aaai2026}  
\usepackage{times}  
\usepackage{helvet}  
\usepackage{courier}  
\usepackage[hyphens]{url}  
\usepackage{graphicx} 
\usepackage{natbib}  
\usepackage{caption} 
\usepackage{algorithm}
\usepackage{algorithmic}

\usepackage{color}
\usepackage{array}
\usepackage{booktabs}
\usepackage{multirow}
\usepackage{graphicx}
\usepackage{microtype}
\usepackage{subfigure}
\usepackage{bbding}
\usepackage{pifont}
\usepackage{mathtools}
\usepackage{dsfont}
\usepackage{colortbl}
\usepackage[switch]{lineno}
\usepackage{epsfig}
\usepackage{tabularx}
\usepackage[utf8]{inputenc}
\usepackage{amssymb}
\usepackage{xspace}
\usepackage{makecell}

\usepackage[capitalize,noabbrev]{cleveref}
\usepackage[T1]{fontenc}

\definecolor{commentcolor}{RGB}{110,154,155}   
\newcommand{\numstd}[2]{#1 {\scriptsize$\pm$#2}}

\newcommand{\mymodel}{DiP\xspace}

\usepackage{newfloat}
\usepackage{listings}
\DeclareCaptionStyle{ruled}{labelfont=normalfont,labelsep=colon,strut=off} 
\floatstyle{ruled}
\newfloat{listing}{tb}{lst}{}
\floatname{listing}{Listing}
\title{Multimodal Graph Representation Learning  with Dynamic Information Pathways}
\author{
    Xiaobin Hong\textsuperscript{\rm 1}, Mingkai Lin\textsuperscript{\rm 1*}, Xiaoli Wang\textsuperscript{\rm 2}, Chaoqun Wang\textsuperscript{\rm 3}, Wenzhong Li\textsuperscript{\rm 1}\thanks{Mingkai Lin and Wenzhong Li are co-corresponding authors.}
}
\affiliations{
    \textsuperscript{\rm 1}State Key Laboratory for Novel Software Technology, Nanjing University\\
    \textsuperscript{\rm 2}Nanjing Forest University, College of Information Science and Technology\\
    \textsuperscript{\rm 3}Zhejiang Provincial Seaport Investment \& Operation Group Co. Ltd\\
    xiaobinhong@smail.nju.edu.cn; mingkai@nju.edu.cn; lwz@nju.edu.cn\\

}

\usepackage{bibentry}

\begin{document}

\maketitle

\begin{abstract}
Multimodal graphs, where nodes contain heterogeneous features such as images and text, are increasingly common in real-world applications. Effectively learning on such graphs requires both adaptive intra-modal message passing and efficient inter-modal aggregation. However, most existing approaches to multimodal graph learning are typically extended from conventional graph neural networks and rely on static structures or dense attention, which limit flexibility and expressive node embedding learning. In this paper, we propose a novel multimodal graph representation learning framework with Dynamic information Pathways (\mymodel). By introducing modality-specific pseudo nodes, \mymodel enables dynamic message routing within each modality via proximity-guided pseudo-node interactions and captures inter-modality dependence through efficient information pathways in a shared state space. This design achieves adaptive, expressive, and sparse message propagation across modalities with linear complexity. We conduct the link prediction and node classification tasks to evaluate performance and carry out full experimental analyses. Extensive experiments across multiple benchmarks demonstrate that \mymodel consistently outperforms baselines. 
\end{abstract}

\section{Introduction}
\label{sec:introduction}
\begin{figure}[t]
    \centering
    \includegraphics[width=\linewidth]{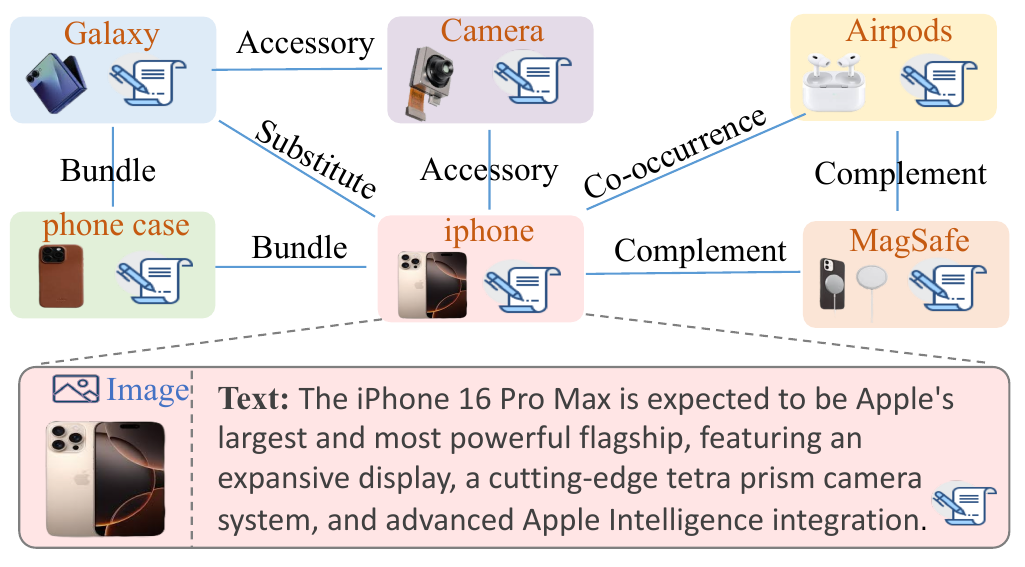}
    \caption{A multimodal ego-graph example from a recommendation system, where its node is attributed with multimodal raw data (i.e., image and text), and the link indicates the complex relations between nodes.}
    \vspace{-4mm}
    \label{fig:example-mmg}
\end{figure}

Graphs are powerful abstractions for modeling relational and structural dependencies across a wide range of domains, including social networks~\cite{zhang2022improving,sharma2024survey}, recommendation systems~\cite{yang2024graph,wang2024distributionally}, and biological interactions~\cite{valous2024graph,ma2023single}. In many real-world applications, graph nodes are enriched with multimodal attributes, such as textual descriptions and visual content, leading to the emergence of Multimodal Graphs (MMGs)~\cite{yan2024graph,zhu2024multimodal}. As illustrated in Figure~\ref{fig:example-mmg}, a typical MMG from a recommendation system represents items as nodes annotated with both images and textual metadata, while edges encode complex semantic or behavioral relationships among them. The inherent heterogeneity of MMGs, characterized by diverse feature modalities and intricate inter-node dependencies, makes them a compelling representation for downstream tasks. such as recommendation~\cite{wei2019mmgcn}, knowledge discovery~\cite{zhang2020graph,zhao2022mose}, and scene understanding~\cite{lee2023vista,wang2025semantic}. To effectively leverage such heterogeneous information, models must not only propagate messages within each modality in a context-aware manner but also enable efficient and meaningful communication across modalities.

While multimodal learning~\citep{zong2024self} has garnered significant attention in recent years, research on multimodal graph representation learning remains relatively nascent. Most existing approaches to multimodal graph learning typically extend conventional graph neural networks (GNNs, like GCN~\cite{kipf2016semi} and GAT~\cite{velivckovic2017graph}) by stacking modality-specific encoders with static graph convolutions~\cite{wei2019mmgcn} or cross-modal attention mechanisms~\cite{tao2020mgat}. While such architectures offer a straightforward extension of unimodal GNNs, they face significant limitations when applied to realistic, large-scale, and semantically complex multimodal graphs. First, there exists a fundamental \textbf{misalignment in information granularity across modalities}. Visual data often encodes fine-grained, instance-level cues such as spatial layout or object parts, whereas textual descriptions tend to abstract high-level semantic concepts. This granularity gap complicates alignment, as a direct fusion of such heterogeneous features often leads to semantic dilution or misinterpretation, especially when applied uniformly across all nodes. Second, most prior methods \textbf{rely on static structure-based aggregation}, either predefined by heuristics or constructed from fixed similarity measures. Such static topology fails to capture dynamic and context-aware dependencies between nodes, and cannot adapt to task-specific signals. Consequently, these rigid structures often lead to known issues such as \textit{over-smoothing}~\cite{oono2019graph} and \textit{over-squashing}~\cite{di2023over,di2023does}. Third, previous strategies exploit \textbf{modal-agnostic fusion}, such as feature concatenation or shared cross-modal attention after modal-independent learning, overlook the complementary nature of different modalities during both local and global aggregation. Without explicitly modeling modality-aware interactions and routing dynamics, these methods fail to fully exploit the expressiveness of MMGs.

To address these challenges, we propose a novel  multimodal graph representation learning method with Dynamic information Pathways (termed \mymodel). \mymodel introduces modality-specific pseudo nodes as lightweight, dynamic intermediaries enabling flexible, efficient, and scalable multimodal graph learning. The core idea is to decouple the complexity of node-level interactions by introducing two information pathways: (i) \textbf{Intra-modal diffusion pathway}: Each modal is equipped with a set of learnable pseudo nodes that mediate the message diffusion through a shared proximity-based attention mechanism. This allows the model to construct flexible message passing pathways that adapt to task-specific context, overcoming the rigidity of static graph structures. (ii) \textbf{Inter-modal aggregation pathway}: Instead of directly modeling dense cross-modal node interactions, \mymodel restricts inter-modality communication to pseudo-to-pseudo interactions. Pseudo nodes from different modalities interact in a shared state space using dynamic proximity, enabling expressive and complementary information fusion at significantly reduced computational cost. This design enables adaptive routing, heterogeneity-aware fusion, and sparse computation in a unified framework. The overall complexity scales linearly with node number, ensuring high scalability. We instantiate \mymodel with a recurrent architecture and shared message update function, improving parameter efficiency and generalizability. As a result, \mymodel supports expressive multimodal reasoning while maintaining linear complexity in graph size, making it suitable for large-scale applications. Experiments on various multimodal graph tasks (i.e., link prediction and node classification) show that \mymodel consistently outperforms existing MMG methods, particularly when modality relationships are complex or dynamically shifting. Our main contributions can be summarized as follows:

\begin{itemize}
    \item We propose \mymodel, a novel framework for multimodal graph representation learning, which enables adaptive, efficient, and scalable message propagation via learnable dynamic information pathways.
    \item We design a multimodal message passing system that constructs dynamic intra- and inter-modality pathways, leading to expressive and context-aware node embeddings.
    \item We conduct extensive experiments on multiple downstream tasks and provide a comprehensive analysis to demonstrate the effectiveness and efficiency of \mymodel.
\end{itemize}

\section{Related Work}
\label{sec:related-work}

\subsection{Multimodal Graph Learning}
\label{sec:multimodal_graph}

Multimodal Graphs (MMGs) indicate that the nodes are associated with multimodal attributes, such as texts and images, which have been widely featured in real-life scenarios. Prior research in multimodal graph learning has largely focused on domain-specific applications such as knowledge graphs~\cite{chen2022hybrid,zeng2023multi}, molecular structures~\cite{jin2018learning}, and brain networks~\cite{wang2023hypergraph}. These models are often tightly coupled with particular tasks and rely on handcrafted designs or domain expertise, which limits their ability to generalize across different graph structures, modalities, or learning objectives. More recently, MMGL~\cite{yoon2023multimodal} attempts to bridge this gap by employing foundation models from multiple modalities on multimodal graphs. However, its focus remains constrained to generative settings, leaving the challenge of building transferable, task-agnostic representations for multimodal graphs largely unaddressed. UniGraph2~\cite{he2025unigraph2} performs large-scale pretraining on visually grounded heterogeneous graphs using modality-specific encoders and MoE fusion. While these methods provide valuable insights into multimodal fusion, they typically use static intra-modality structures or dense fusion strategies, lacking the adaptability and efficiency necessary for dynamic multimodal graphs.

\subsection{Graph Message Passing}
\label{sec:graph_mp}

Message passing is the fundamental paradigm of GNNs, which aggregates the information from neighbors and updates node states~\cite{kipf2016semi,hamilton2017inductive,hong2024label}. Many graph learning methods enhance expressiveness by approximating filters with parameterized polynomials~\cite{chien2020adaptive,gasteiger2019diffusion}, but are limited to low-order terms due to computational cost, leading to local aggregation and issues like over-smoothing~\cite{li2018deeper,oono2019graph,lin2024scalable} and over-squashing~\cite{alon2020bottleneck,topping2021understanding}. To address these problems, recent works introduce auxiliary structures—such as edge shortcuts~\cite{hong2021graph,gutteridge2023drew}, graph pooling~\cite{gao2019graph,ranjan2020asap,hong2021variational}, and pseudo nodes~\cite{liu2022boosting,shirzad2023exphormer} to decouple message passing from the input topology. While edge shortcut methods dynamically rewire graphs for improved multi-hop communication, pseudo nodes offer global context but are often implemented with static, uniform connections that hinder adaptive message propagation~\cite{shirzad2023exphormer}. In this work, we revisit pseudo nodes with a dynamic and task-aware design that promotes efficient and adaptive global communication.


\begin{figure*}[ht]
    \centering
    \includegraphics[width=0.96\linewidth]{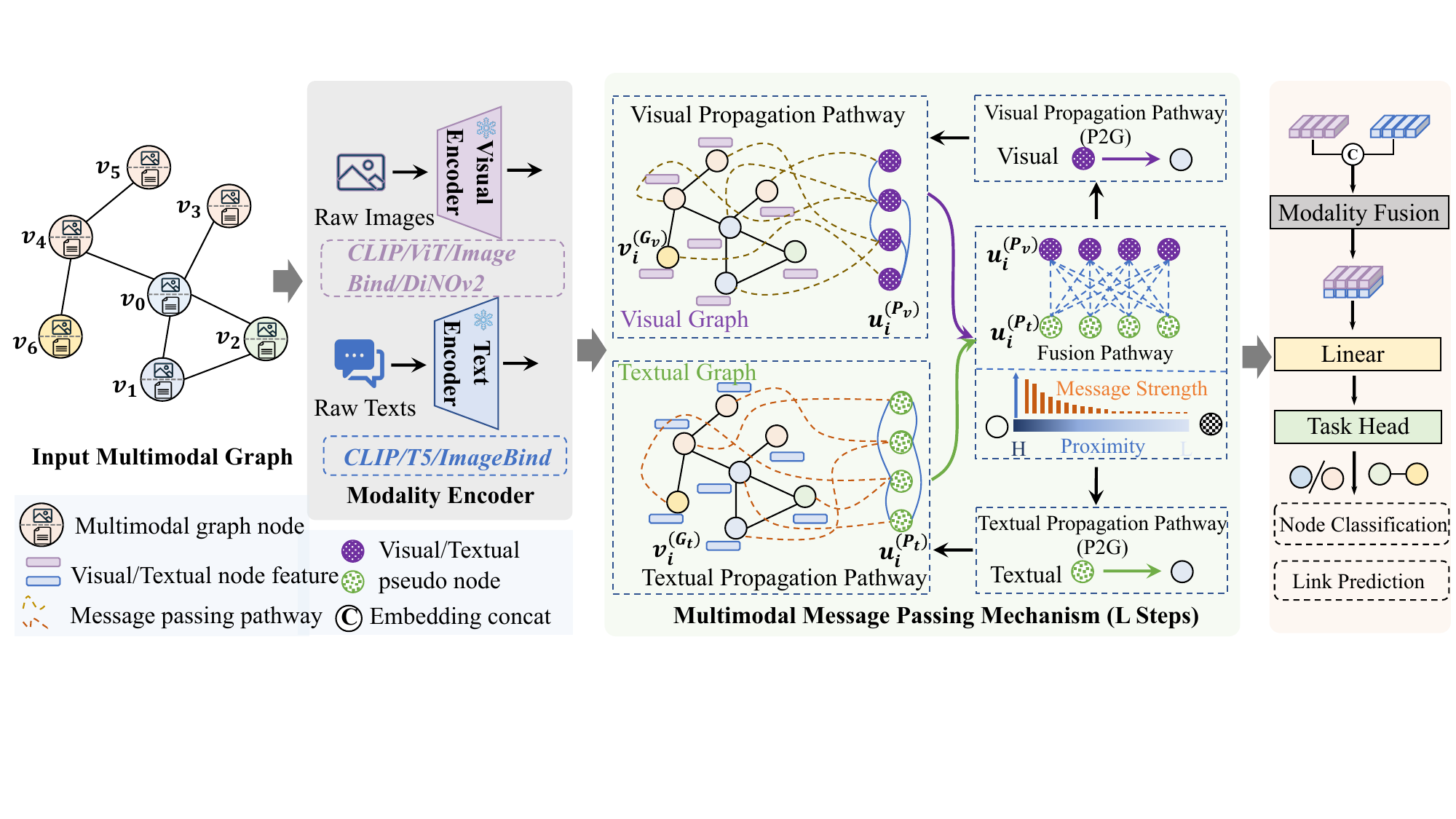}
    \caption{The overview framework of \mymodel. \mymodel first encodes the raw images and texts with frozen modality encoders for multimodal graph nodes. The recursive $L$-steps multimodal message passing mechanism is the key component of \mymodel, which consists of the \textit{Intra-Modal Diffusion Pathway} and \textit{Inter-Modal Aggregation Pathway} modules, which output expressiveness node representations incorporating the intra- and inter-modal message passing. Finally, the learned multimodal embeddings are fed to the task heads for link prediction and node classification training.}
    \label{fig:framework}
\end{figure*}

\section{Methodology}
\label{sec:method}

As shown in Figure~\ref{fig:framework}, our \mymodel uses the frozen modality encoders to embed the raw modality data, followed by $L$-step multimodal message passing that models adaptive intra- and inter-modal pathways through measurable node relations in a shared space. The resulting node embeddings are used for downstream tasks. We first define the dynamic pathway construction, where modality-specific pseudo nodes reduce computational cost and enable cross-modal and cross-neighborhood communication. We then detail how \mymodel learns node displacements to facilitate adaptive message passing.

\subsection{Preliminaries}
\label{sec:preliminaries}


\textbf{Notations.} We define a multimodal graph as $\mathcal{G}=(\mathcal{V}, \mathcal{E}, \mathrm{X})$, where $\mathcal{V}=\{v_1,\cdots,v_n\}$ is the node set with multimodal raw data. Following the benchmark~\cite{zhu2025mosaic}, we focus on image-text pair graphs, denoting visual and text nodes as $\mathcal{V}^{(v)}$ and $\mathcal{V}^{(t)}$. $\mathcal{E}=\{e_{v_i,v_j}|v_j\in\mathcal{N}(v_i)\}$ represents edges, with $\mathcal{N}(v_i)$ the one-hop neighbors of $v_i$. $\mathrm{X}=[\mathrm{X}^{(v)},\mathrm{X}^{(t)}]=\{\mathrm{x}_{v_1},\cdots,\mathrm{x}_{v_n}\}$ denotes multimodal node attributes. For each modality, we introduce pseudo nodes $\mathcal{P}^{(v)}$ and $\mathcal{P}^{(t)}$ with tunable sizes $n_{p_v}$ and $n_{p_t}$ to enable adaptive multimodal message passing.

\noindent \textbf{Modality Encoder.} We employ modality encoders to project raw images and texts into latent spaces, yielding visual features $\mathrm{X}^{(v)} \in \mathbb{R}^{n \times d_v}$ and textual features $\mathrm{X}^{(t)} \in \mathbb{R}^{n \times d_t}$. For the visual encoder $E_v$, we adopt CLIP~\cite{radford2021learning}, ViT~\cite{dosovitskiy2020image}, DINOv2~\cite{oquab2023dinov2}, and ImageBind~\cite{girdhar2023imagebind}, covering supervised, self-supervised, and joint vision-language paradigms. ImageBind further supports embedding diverse modalities (e.g., audio, video) into a shared space for generalized multimodal learning. For the text encoder $E_t$, we use CLIP, T5~\cite{raffel2020exploring}, and ImageBind.

\subsection{Dynamic Pathway Construction}
\label{sec:pathway}

Conventional GNNs rely on static graph structures, limiting message passing to local neighborhoods. To enable global interactions, some works introduce pseudo nodes as intermediaries; however, they usually learn pairwise edge weights between nodes and pseudo nodes~\cite{gilmer2017neural, liu2022boosting}, leading to parameter growth linear in graph size. To enhance scalability, we draw inspiration from DyN~\cite{pei2023dynamics} and $\text{N}^2$~\cite{sun2024towards}, which model neuronal interactions via spatially parameterized functional connections instead of individualized edge weights. These methods improve parameter efficiency using a shared path integral function conditioned on spatial coordinates, allowing dynamic modulation of information flow. Following this analogy, we treat graph nodes as neurons and introduce a unified state space $\mathcal{S}\in\mathbb{R}^{d_s}$, where each node and pseudo node has a learnable state embedding encoding both modality-specific features and local topology. A shared metric function over $\mathcal{S}$ computes proximity between nodes and pseudo nodes, constructing dynamic pathways without per-edge parameters.

Given a pseudo node set $\mathcal{P}=\{u_1,\cdots,u_p\}$\footnote{For simplicity, modal subscripts are omitted.}, we embed them in $\mathcal{S}$ as learnable parameters $\mathrm{H}=\{\mathrm{h}_{u_1},\cdots,\mathrm{h}_{u_p}\}^\top\in\mathbb{R}^{n_p\times d_s}$. These pseudo nodes interact with modality graph nodes for adaptive message passing through dynamic pathways. Encoded node features are projected into the common space $\mathcal{S}$ as $\mathrm{Z}=\{\mathrm{z}_{v_1},\cdots,\mathrm{z}_{v_n}\}=f(\mathrm{X})\in\mathbb{R}^{n\times d_s}$, where $f:\mathbb{R}^{d}\mapsto\mathcal{S}$ is permutation equivariant. To capture complex relations, we approximate non-linear interactions with a multi-channel path integral, projecting each node into $\tau$ channels for proximity computation, analogous to multi-head attention~\cite{vaswani2017attention}:
\begin{align}
    \phi(\mathrm{v}_{i}, \mathrm{v}_{j}) = \sum_{t=1}^{\tau}\lambda_t\, \mathrm{z}_{tv_i}^{\top}\mathrm{z}_{tv_j}, \quad \mathrm{z}_{tv} = \sigma(\mathrm{z}_{t}),
    \label{eq:proximity}
\end{align}
where $\lambda_t$ is a learnable weight, $\sigma(\cdot)$ is a non-linear function with linear mapping followed by $\mathtt{LeakyReLU}(\cdot)$, and $\tau$ is the channel scale (following $\text{N}^2$, we set $\tau=8$).

\subsection{Multimodal Message Passing Mechanism}
\label{sec:multimodal_mp}
We propose a Multimodal Message Passing Mechanism that jointly models intra-modal diffusion and inter-modal aggregation for effective node representation learning. As shown in Figure~\ref{fig:framework}, our framework consists of two core components: the \textbf{Intra-Modal Diffusion Pathway} and the \textbf{Inter-Modal Aggregation Pathway}. Within each modality, we introduce pseudo nodes as latent mediators to enable scalable and expressive message propagation. Specifically, the intra-modal pathway includes both graph-to-pseudo (G2P) and pseudo-to-graph (P2G) routes, allowing pseudo nodes to gather and redistribute information across the entire graph. For inter-modal interaction, modality-specific pseudo nodes act as bridges that exchange information between different modalities. The updated pseudo node representations are then propagated back to their respective graph nodes to refine the node embeddings in a modality-aware and adaptive manner. 
For clarity, we omit modality-specific subscripts in intra-modal formulations, as the visual and textual branches adopt symmetric architectures.

\textbf{Intra-Modal Diffusion Pathway (G2P).} This pathway aims to capture the visual/text in-modality patterns, where pseudo nodes serve as global proxies. Following the common practice in GNNs~\cite{kipf2016semi}, we interpret the interaction between embedded nodes as message passing. To achieve this, both graph node $v$ and pseudo node $u$ learn their message $\mathrm{m}_{v}, \mathrm{m}_{u} \in \mathbb{R}^{d}$ to be passed in the state space $\mathcal{S}$.The messages of graph nodes are initialized with node features, and learnable parameters construct pseudo nodes. We first perform local message passing ($\mathtt{LocalMP}(\mathrm{M}, \mathcal{E})$) for graph nodes by exchanging their message among the ego-neighborhood:
\begin{align}
    \label{eq:local_mp}
    \mathrm{m}_{v} = \psi(\mathrm{z}_{v}, \frac{1}{|\mathcal{N}(v)|}\sum_{v_j\in \mathcal{N}(v)}\mathrm{z}_{v_j}),
\end{align}
where $\psi(\cdot)$ denotes message aggregation function. The topology-coupled message passing encodes the local structure into node states, and we then use the pseudo nodes to aggregate the global modality patterns. Given graph node states $\mathrm{Z} \in \mathbb{R}^{n\times d_s}$, the pseudo node states $\mathrm{H} \in \mathbb{R}^{n_p \times d_s}$, and graph node message $\mathrm{M}^{G}=(\mathrm{m}_{v_1}, \cdots, \mathrm{m}_{v_n})^{\top}\in \mathbb{R}^{n\times d}$, the global message-passing process can be formulated as:
\begin{align}
    &\mathrm{D} = \mathrm{W}^{GP}\mathrm{M}^{G},\quad \mathrm{W}^{GP}_{ij}=\phi(\mathrm{H}_{i,\cdot}, \mathrm{Z}_{j,\cdot}), \label{eq:global_diffuse} \\
    &\mathrm{\hat{D}} = \mathrm{W}^{PP}\mathrm{D}, \Delta \mathrm{H} = \sigma(\mathrm{\hat{D}}), \quad \mathrm{W}^{PP}_{ij}=\phi(\mathrm{H}_{i,\cdot}, \mathrm{H}_{j,\cdot}), \label{eq:global_refine} \\
    &\mathrm{\hat{M}}^{G} = \mathrm{W}^{PG}\sigma(\mathrm{\hat{D}}),\mathrm{W}^{PG}_{i,j}=\phi(\mathrm{Z}_{i,\cdot}, [\mathrm{H}+\Delta \mathrm{H}]_{j,\cdot}), \label{eq:global_agg}
\end{align}
where $\mathrm{W}^{GP} \in \mathbb{R}^{n_p \times n}$ denotes the edge weight matrix form graph nodes to pseudo nodes calculated by Eq.~\ref{eq:proximity}, $\mathrm{W}^{PP}$ and $\mathrm{W}^{PG}$ follow the similar definition. The Eq.~\ref{eq:global_diffuse}, \ref{eq:global_refine}, and \ref{eq:global_agg} perform the message diffusion, refinement, and aggregation process. We compile them as global message passing function($\mathtt{GlobMP}(\cdot)$). Note that the complexity of our global message passing in $\mathcal{O}(\tau nn_p)$, with $\tau n_p\ll n$, significantly lower than $\mathcal{O}(n^2)$ in the dense scenario. 

Intra-modality G2P pathway performs both local and global message passing. At the local level, graph nodes aggregate their message and update the node states:
\begin{align}
    \label{eq:local_g2p}
    \mathrm{\hat{Z}}^{(l)} = \mathrm{Z}^{(l-1)} + \sigma(\mathrm{M}^{(l)}), \mathrm{M}^{(l)} = \mathtt{LocalMP}(\mathrm{M}||\mathrm{Z}, \mathcal{E}).
\end{align}
At the global level, we update the node states and messages according to global modality patterns:

\begin{equation}
    \label{eq:global_g2p}
    \begin{aligned}
        &\mathrm{\hat{M}}^{G},\Delta \mathrm{H}=\mathtt{GlobMP}(\mathrm{H}, \mathrm{M}^{G}, \mathrm{Z}),~ \mathrm{\tilde{Z}}^{(l)} = \mathrm{\hat{Z}}^{(l)} + \sigma(\mathrm{\hat{M}}^{G}), \\
        &\mathrm{\tilde{M}}^{(l)} = \mathrm{M}^{(l-1)}+ \mathrm{\hat{M}}^{G},~\mathrm{\hat{H}}^{(l)}=\mathrm{H}^{(l-1)}+\Delta \mathrm{H}.
    \end{aligned}
\end{equation}

\textbf{Inter-Modal Aggregation Pathway} aims to enable inter-modality message passing by constructing cross-modal interactions. Given the updated visual and text modality pseudo node states $\mathrm{\hat{H}}^{(l)}_{v}\in \mathbb{R}^{n_{p_v} \times d_s}, \mathrm{\hat{H}}^{(l)}_{t}\in \mathbb{R}^{n_{p_t} \times d_s}$ from the intra-modality pathway, we use the proximity measurement to construct the adaptive message pathway for capturing cross-modal patterns:
\begin{align}
    \label{eq:inter_modal}
    \mathrm{\tilde{H}}^{(l)}_{v} &= \mathrm{\hat{H}}^{(l)}_{v} + \mathrm{W}^{tv}\mathrm{\hat{H}}^{(l)}_{t}, \quad \mathrm{W}^{tv} = \phi(\mathrm{\hat{H}}^{(l)}_{t}, \mathrm{\hat{H}}^{(l)}_{v}), \\
    \mathrm{\tilde{H}}^{(l)}_{t} &= \mathrm{\hat{H}}^{(l)}_{t} + \mathrm{W}^{vt}\mathrm{\hat{H}}^{(l)}_{v}, \quad \mathrm{W}^{vt} = \phi(\mathrm{\hat{H}}^{(l)}_{v}, \mathrm{\hat{H}}^{(l)}_{t}).
\end{align}

\begin{table*}[t]
    \centering
    \resizebox{\textwidth}{!}{
    \begin{tabular}{@{}ll c ccc c ccc c ccc@{}}
    \toprule
       \multirow{2}{*}{\scalebox{1.2}{Encoder}}  & \multirow{2}{*}{\scalebox{1.2}{Model}} && \multicolumn{3}{c}{Amazon-Sports} &&  \multicolumn{3}{c}{Amazon-Cloth} && \multicolumn{3}{c}{Goodreads-LP} \\
       
       \cmidrule(lr){4-6} \cmidrule(lr){8-10} \cmidrule(lr){12-14}
       & && \textbf{MRR$\uparrow$} & \textbf{Hit$@ 1\uparrow$} & \textbf{Hit$@ 10\uparrow$} && \textbf{MRR$\uparrow$} & \textbf{Hit$@ 1\uparrow$} & \textbf{Hit$@ 10\uparrow$} && \textbf{MRR$\uparrow$} & \textbf{Hit$@ 1\uparrow$} & \textbf{Hit$@ 10\uparrow$} \\
    \midrule
       \multirow{8}{*}{\parbox{1.8cm}{\raggedright $E_v = \text{CLIP}$ \\ $E_t = \text{CLIP}$}}
       & MLP && \numstd{28.22}{0.09} & \numstd{14.54}{0.16} & \numstd{59.40}{0.08} && \numstd{21.10}{0.04} & \numstd{10.70}{0.03} & \numstd{42.77}{0.05} && \numstd{11.03}{0.06} & \numstd{4.87}{0.04} & \numstd{21.61}{0.11} \\
       
       & GCN & & \numstd{31.38}{0.08} & \numstd{16.58}{0.13} & \numstd{66.14}{0.08} & & \numstd{22.28}{0.05} & \numstd{11.83}{0.04} & \numstd{43.52}{0.10} & & \numstd{25.34}{0.06} & \numstd{13.81}{0.12}& \numstd{50.36}{0.14} \\
       
       & SAGE & & \numstd{33.83}{0.08} & \numstd{17.57}{0.14} & \numstd{71.90}{0.07} & & \numstd{24.58}{0.18} & \numstd{12.16}{0.11} & \numstd{51.12}{0.09} & & \numstd{44.10}{1.37} & \numstd{32.32}{1.38} & \numstd{69.07}{1.19}\\
       
       & BUDDY & & \numstd{31.55}{0.13}& \numstd{15.05}{0.43}& \numstd{70.92}{0.25} & & \numstd{23.44}{0.26} & \numstd{11.06}{0.20} & \numstd{51.08}{0.50} & & \numstd{43.25}{0.23} & \numstd{31.84}{0.35} & \numstd{67.93}{0.03} \\

       & MMGCN & & \numstd{31.96}{0.10} & \numstd{16.35}{0.11} & \numstd{68.46}{0.08} & & \numstd{22.20}{0.05} & \numstd{10.76}{0.1} & \numstd{46.62}{0.12} & & \numstd{31.84}{0.09} & \numstd{18.63}{0.31} & \numstd{59.85}{0.19} \\ 

       & MGAT & & \numstd{27.56}{0.30} & \numstd{13.55}{0.29} & \numstd{60.21}{0.21} & & \numstd{21.38}{0.23} & \numstd{10.39}{0.22} & \numstd{44.60}{0.36} & & \numstd{44.75}{1.23} & \textbf{\numstd{34.53}{1.48}} & \numstd{62.81}{0.64} \\

       & UniGraph2 && \numstd{31.61}{0.14} & \numstd{15.72}{0.28} & \numstd{65.52}{0.36} & & \numstd{23.58}{0.63} & \numstd{12.54}{0.26} & \numstd{49.65}{0.28} & & \numstd{29.68}{0.52} & \numstd{21.75}{0.22} & \numstd{58.71}{0.36} \\

       & \mymodel(ours) & & \textbf{\numstd{34.26}{0.31}} & \textbf{\numstd{18.45}{0.27}} & \textbf{\numstd{72.54}{0.71}} && \textbf{\numstd{25.19}{0.06}} & \textbf{\numstd{14.26}{0.17}} & \textbf{\numstd{52.09}{0.35}} && \textbf{\numstd{45.13}{0.24}} & \numstd{34.47}{0.28} & \textbf{\numstd{70.49}{0.62}}\\
    \midrule

        \multirow{8}{*}{\parbox{1.8cm}{\raggedright $E_v = \text{ViT}$ \\ $E_t = \text{T5}$}}
        & MLP && \numstd{24.81}{0.05} & \numstd{11.63}{0.05} & \numstd{54.78}{0.04}& & \numstd{17.65}{0.06} & \numstd{8.14}{0.04} & \numstd{36.77}{0.06} & & \numstd{11.10}{0.17} & \numstd{4.84}{0.15} & \numstd{21.94}{0.24} \\

        & GCN && \numstd{30.83}{0.07} & \numstd{16.31}{0.08} & \numstd{64.76}{0.15}& & \numstd{21.60}{0.05} & \numstd{11.37}{0.03} & \numstd{42.29}{0.14} & & \numstd{26.50}{0.10} & \numstd{14.86}{0.08} & \numstd{51.54}{0.14}\\

        & SAGE && \numstd{32.01}{0.10} & \numstd{15.94}{0.17} & \numstd{69.84}{0.21} & & \numstd{23.11}{0.05} & \numstd{11.10}{0.04} & \numstd{48.89}{0.09} & & \numstd{44.79}{0.18} & \numstd{33.11}{0.21} & \numstd{69.43}{0.18}\\

        & BUDDY && \numstd{30.41}{0.40} & \numstd{14.11}{0.28} & \numstd{69.55}{0.80} & & \numstd{22.82}{0.19} & \numstd{10.24}{0.12} & \numstd{51.04}{0.39} & & \numstd{43.18}{0.53} & \numstd{31.73}{0.54} & \numstd{67.89}{0.57} \\ 

        & MMGCN && \numstd{30.33}{0.03} & \numstd{15.01}{0.05} & \numstd{66.41}{0.11} & & \numstd{19.45}{0.34} & \numstd{9.22}{0.20} & \numstd{40.49}{0.61}& & \numstd{31.11}{0.25} & \numstd{19.30}{0.45} & \numstd{56.24}{0.19} \\
        & MGAT && \numstd{30.15}{0.34} & \numstd{15.28}{0.34} & \numstd{64.84}{0.41} & & \numstd{20.59}{0.41} & \numstd{9.79}{0.30} & \numstd{43.44}{0.76} & & \numstd{35.26}{1.21} & \numstd{35.23}{1.62} & \numstd{62.90}{1.89} \\

        & UniGraph2 && \numstd{32.17}{0.09} & \numstd{16.80}{0.14} & \numstd{67.81}{0.29} & & \numstd{22.68}{0.72} & \numstd{12.03}{0.20} & \numstd{48.17}{0.25} & & \numstd{28.52}{0.08} & \numstd{22.01}{0.29} & \numstd{57.13}{0.17} \\

        & \mymodel(ours) && \textbf{\numstd{33.74}{0.15}} & \textbf{\numstd{18.21}{0.09}} & \textbf{\numstd{70.53}{0.26}} && \textbf{\numstd{25.16}{0.18}} & \textbf{\numstd{13.05}{0.17}} & \textbf{\numstd{51.86}{0.63}} && \textbf{\numstd{45.18}{0.24}} & \textbf{\numstd{35.29}{0.34}} & \textbf{\numstd{70.34}{0.23}}\\
    \midrule

        \multirow{8}{*}{\parbox{2.5cm}{\raggedright $E_v = \text{ImageBind}$ \\ $E_t = \text{ImageBind}$}} 
        & MLP && \numstd{30.45}{0.14} & \numstd{15.91}{0.10} & \numstd{64.10}{0.07} & & \numstd{22.18}{0.02} & \numstd{11.42}{0.04} & \numstd{44.86}{0.06} && \numstd{7.73}{0.06} & \numstd{3.37}{0.07} & \numstd{13.26}{0.03} \\
        & GCN && \numstd{31.67}{0.09} & \numstd{17.07}{0.14} & \numstd{65.61}{0.10} & & \numstd{22.81}{0.03} & \numstd{12.27}{0.05} & \numstd{44.28}{0.09} && \numstd{27.56}{1.26} & \numstd{14.31}{1.37} & \numstd{57.25}{0.52} \\
        & SAGE && \numstd{34.32}{0.11} & \numstd{17.87}{0.23} & \numstd{73.04}{0.15} && \numstd{25.20}{0.09} & \numstd{12.63}{0.05} & \numstd{52.53}{0.21} && \numstd{34.61}{0.43} & \numstd{23.82}{0.51} & \numstd{56.67}{0.21} \\
        & BUDDY && \numstd{33.02}{0.44} & \numstd{17.61}{0.43} & \numstd{69.17}{0.43} && \numstd{24.35}{0.24} & \numstd{12.05}{0.46} & \numstd{51.44}{0.87} && \numstd{41.56}{0.61} & \numstd{29.89}{0.91} & \numstd{67.41}{0.05} \\
        & MMGCN && \numstd{31.74}{0.21} & \numstd{16.45}{0.13} & \numstd{67.39}{0.74} && \numstd{24.72}{0.19} & \numstd{12.47}{0.09} & \numstd{51.32}{0.56} && \numstd{26.32}{0.23} & \numstd{16.05}{0.22} & \numstd{46.37}{0.66} \\
        & MGAT && \numstd{30.15}{0.12} & \numstd{15.50}{0.05} & \numstd{64.20}{0.43} && \numstd{22.13}{0.27} & \numstd{10.96}{0.15} & \numstd{45.84}{0.57} && \numstd{34.77}{0.49} & \numstd{34.95}{0.61} & \numstd{62.51}{0.47} \\
        & UniGraph2 && \numstd{32.35}{0.06} & \numstd{15.28}{0.30} & \numstd{67.83}{0.15} & & \numstd{24.37}{0.29} & \numstd{11.75}{0.23} & \numstd{50.21}{0.06} & & \numstd{32.43}{0.57} & \numstd{26.55}{0.21} & \numstd{59.77}{0.38} \\
        & \mymodel(ours) && \textbf{\numstd{35.16}{0.53}} & \textbf{\numstd{19.57}{0.14}} & \textbf{\numstd{74.23}{0.27}} && \textbf{\numstd{26.18}{0.19}} & \textbf{\numstd{14.08}{0.61}} & \textbf{\numstd{54.17}{0.24}} && \textbf{\numstd{46.13}{0.12}} & \textbf{\numstd{36.07}{0.24}} & \textbf{\numstd{71.06}{0.04}}\\
    \midrule

        \multirow{8}{*}{\parbox{2.5cm}{\raggedright $E_v = \text{DINOv2}$ \\ $E_t = \text{T5}$}} 
        & MLP && \numstd{24.81}{0.16} & \numstd{11.62}{0.18} & \numstd{54.97}{0.22} && \numstd{17.53}{0.11} & \numstd{8.07}{0.09} & \numstd{36.53}{0.26} &&\numstd{10.28}{0.04} & \numstd{4.49}{0.05} & \numstd{19.86}{0.03} \\
        & GCN && \numstd{30.42}{0.02} & \numstd{16.02}{0.03} & \numstd{64.02}{0.06} && \numstd{21.19}{0.08} & \numstd{11.09}{0.06} & \numstd{41.46}{0.16} && \numstd{28.21}{1.12} & \numstd{15.11}{1.06} & \numstd{57.94}{0.95}\\
        & SAGE && \numstd{32.20}{0.12} & \numstd{16.19}{0.20} & \numstd{69.98}{0.32} && \numstd{22.98}{0.01} & \numstd{11.12}{0.04} & \numstd{48.28}{0.11} && \textbf{\numstd{45.61}{0.22}} & \numstd{34.01}{0.27} & \numstd{70.01}{0.11}\\
        & BUDDY && \numstd{30.02}{0.34}& \numstd{13.78}{0.19} & \numstd{69.18}{0.67} &&\numstd{22.95}{0.06} & \numstd{10.45}{0.09} & \numstd{50.87}{0.61} && \numstd{43.25}{0.13} & \numstd{31.77}{0.33} & \numstd{68.08}{0.42}\\
        & MMGCN && \numstd{30.04}{0.27} & \numstd{14.98}{0.07} & \numstd{64.56}{0.56} && \numstd{21.77}{0.23} & \numstd{10.47}{0.12} & \numstd{45.81}{0.52} && \numstd{27.64}{0.95} & \numstd{16.21}{0.65} & \numstd{51.46}{1.71} \\
        & MGAT && \numstd{28.91}{0.09} & \numstd{14.47}{0.18} & \numstd{62.11}{0.22} & & \numstd{21.42}{0.13} & \numstd{10.38}{0.13} & \numstd{44.11}{0.50} & & \numstd{74.89}{1.46} & \numstd{64.70}{1.98} & \numstd{92.92}{0.41}\\
        & UniGraph2 && \numstd{31.08}{0.16} & \numstd{16.04}{0.32} & \numstd{66.73}{0.30} & & \numstd{22.86}{0.15} & \numstd{12.14}{0.06} & \numstd{49.78}{0.42} & & \numstd{29.74}{0.62} & \numstd{22.68}{0.18} & \numstd{59.45}{0.28} \\
        & \mymodel(ours) && \textbf{\numstd{33.47}{0.25}} & \textbf{\numstd{17.93}{0.12}} & \textbf{\numstd{70.42}{0.06}} && \textbf{\numstd{24.35}{0.19}} & \textbf{\numstd{13.42}{0.26}} & \textbf{\numstd{51.44}{0.08}} && \numstd{44.76}{0.27} & \textbf{\numstd{34.75}{0.65}} & \textbf{\numstd{71.42}{0.28}} \\
    \bottomrule
    \end{tabular}
    }
    \caption{Link prediction results on Amazon-Sports, Amazon-Cloth, and Goodreads-LP. \textbf{Bold} is the best result.}
    \label{tab:lp-results}
\end{table*}

\textbf{Intra-Modal Diffusion Pathway (P2G)} propagates the updated pseudo node states from P2P pathway into the graph nodes through global message passing across modalities. This enables the in-modal node states to perceive cross-modal information and enhances the expressive power of node embedding. Given the pseudo node states $\mathrm{\tilde{H}}^{(l)}\in \mathbb{R}^{n_p \times d_s}$, the graph node states $\mathrm{\tilde{Z}}^{(l)}\in \mathbb{R}^{n\times d_s}$, and node messages $\mathrm{\tilde{M}}^{(l)} \in \mathbb{R}^{n\times d}$, we perform adaptive global message passing:
\begin{equation}
    \label{eq:global_p2g}
    \begin{aligned}
        &\mathrm{M}^{G}, \Delta \mathrm{H} = \mathtt{GlobMP}(\mathrm{\tilde{H}}^{(l)}, \mathrm{\tilde{M}}^{(l)}, \mathrm{\tilde{Z}}^{(l)}), ~\mathrm{Z}^{(l)}=\mathrm{\tilde{Z}}^{(l)}+\sigma(\mathrm{M}^{G}),\\
        &\mathrm{M}^{(l)} = \mathrm{\tilde{M}}^{(l)} + \mathrm{M}^{G}, \quad \mathrm{H}^{(l)}=\mathrm{\tilde{H}}^{(l)} + \Delta \mathrm{H}.
    \end{aligned}
\end{equation}

\mymodel updates the states of the embedded nodes recursively with a single recurrent layer, and the associated parameters are shared across steps. After recursive $L$-steps multimodal message passing system, the graph nodes receive their final states $\mathrm{Z}^{(L)}_{v}, \mathrm{Z}^{(L)}_{t} \in \mathbb{R}^{n\times d_s}$ for visual and text modality. The learned node states are then sent to the downstream task heads for objective output.

\subsection{Modality Fusion and Task Head}
\label{sec:fusion_task}
This module receives the final-layer representations from different modalities and integrates them for downstream tasks. Specifically, we perform modality fusion via straightforward concatenation followed by a linear projection:
\begin{align}
    \label{eq:concat}
    \mathrm{Z}^{(L)} = g(\mathtt{Concat}[\mathrm{Z}^{(L)}_{v}, \mathrm{Z}^{(L)}_{t}]),
\end{align}
where $g: \mathbb{R}^{2d_s} \mapsto \mathbb{R}^{d}$ is a linear projector. 
In this study, we evaluate \mymodel in multimodal graph link prediction and node classification tasks.
We attach a lightweight task head atop the fused representation $\mathrm{Z}^{(L)}$ to support different learning objectives. For node classification, we apply a softmax classifier (implemented by a 2-layer MLP) to predict the label of each node based on its embedding. For link prediction, we estimate the likelihood of an edge between two nodes using the inner product of their embeddings, followed by a sigmoid activation. These tasks are jointly used to assess the representational quality and generalization ability of our model in multimodal graph scenarios.
\section{Experiments}
\label{sec:experiment}

\subsection{Experiments Setup}
\label{sec:setup}

We evaluate \mymodel on two fundamental graph tasks (i.e., link prediction and node classification) in comparison with a topology-free method (i.e., MLP), three conventional GNNs (i.e., GCN~\cite{kipf2016semi}, SAGE~\cite{hamilton2017inductive}, and BUDDY~\cite{chamberlain2022graph}), and three multimodal GNNs (i.e., MMGCN~\cite{wei2019mmgcn}, MGAT~\cite{tao2020mgat}, and UniGraph2~\cite{he2025unigraph2}). Experiments are conducted on five real-world multimodal graph datasets (Three for link prediction, i.e., Amozon-Sports, Amazon-Cloth, and Goodreads-LP. Two for node classification, i.e., Ele-Fashion, Goodreads-NC), which are proposed by \texttt{MM-GRAPH}~\cite{zhu2025mosaic}. For evaluation metrics, we report Mean Reciprocal Rank (MRR), Hits@10, and Hits@1, the three most common-used evaluation metrics for link prediction and accuracy for node classification~\cite{li2023evaluating}. We implement \mymodel using PyTorch 2.4.0 and CUDA 12.2. All experiments are conducted on 4*Tesla V100-SXM2-32GB GPUs. 

\subsection{Multimodal Graph Link Prediction}
\label{sec:exp_lp}

\begin{table*}[th]
    \centering
    \resizebox{\textwidth}{!}{
    \begin{tabular}{l c ll c ccccccc}
    \toprule
        Dataset && $E_v$ & $E_t$ && MLP & GCN & SAGE & MMGCN & MMGAT & UniGraph2 & \mymodel(ours) \\
    \midrule
        \multirow{4}{*}{Ele-Fashion} && CLIP & CLIP && \numstd{85.16}{0.03} & \numstd{79.83}{0.03} & \numstd{87.10}{0.02} & \numstd{86.10}{0.50} & \numstd{84.66}{0.29} & \numstd{85.93}{0.15} & \textbf{\numstd{87.93}{0.04}} \\
        && ViT & T5 &&  \numstd{84.98}{0.05} & \numstd{79.63}{0.07} & \numstd{84.41}{0.09} & \numstd{82.39}{0.30} & \numstd{84.01}{0.08} & \numstd{85.47}{0.20} & \textbf{\numstd{87.19}{0.06}}\\
        && ImageBind & ImageBind && \numstd{88.73}{0.01} & \numstd{80.35}{0.02} & \numstd{87.71}{0.13} & \numstd{86.21}{0.84} & \numstd{86.12}{0.08} & \numstd{86.52}{0.31} & \textbf{\numstd{89.50}{0.07}}\\
        && DINOv2 & T5 && \numstd{84.87}{0.01} & \numstd{79.37}{0.04} & \numstd{85.31}{0.09} & \numstd{85.53}{0.33} & \numstd{84.54}{0.27} & \numstd{86.18}{0.09} & \textbf{\numstd{87.46}{0.05}}\\
    \midrule
        \multirow{4}{*}{Goodreads-NC} && CLIP & CLIP &&  \numstd{72.29}{0.02} & \numstd{81.61}{0.01} & \numstd{83.30}{0.02} & \numstd{83.29}{0.20} & \numstd{76.48}{0.59} & \numstd{82.56}{0.35} & \textbf{\numstd{85.37}{0.12}}\\
        && ViT & T5 && \numstd{67.82}{0.02} & \numstd{81.67}{0.01} & \numstd{83.30}{0.02} & \numstd{81.85}{0.22} & \numstd{75.43}{0.76} & \numstd{81.93}{0.41} & \textbf{\numstd{86.64}{0.07}}\\
        && ImageBind & ImageBind && \numstd{58.75}{0.05} & \numstd{78.91}{0.04} & \numstd{80.39}{0.21} & \numstd{80.58}{1.08} & \numstd{69.45}{6.25} & \numstd{80.52}{0.08} & \textbf{\numstd{83.21}{0.13}}\\
        && DINOv2 & T5 && \numstd{68.83}{0.03} & \numstd{81.71}{0.03} & \numstd{82.99}{0.08} & \numstd{82.44}{0.11} & \numstd{74.98}{1.23} & \numstd{81.63}{0.21} & \textbf{\numstd{84.32}{0.15}}\\
    \bottomrule
    \end{tabular}
    }
    \caption{Node classification results on Goodreads-NC and Ele-Fashion. \textbf{Bold} indicates the best performance.}
    \label{tab:nc-results}
\end{table*}


\begin{figure*}[t]
    \centering
    \subfigure[Over-smoothing effects]{
        \label{fig:over_smoothing}
        \includegraphics[width=0.31\linewidth]{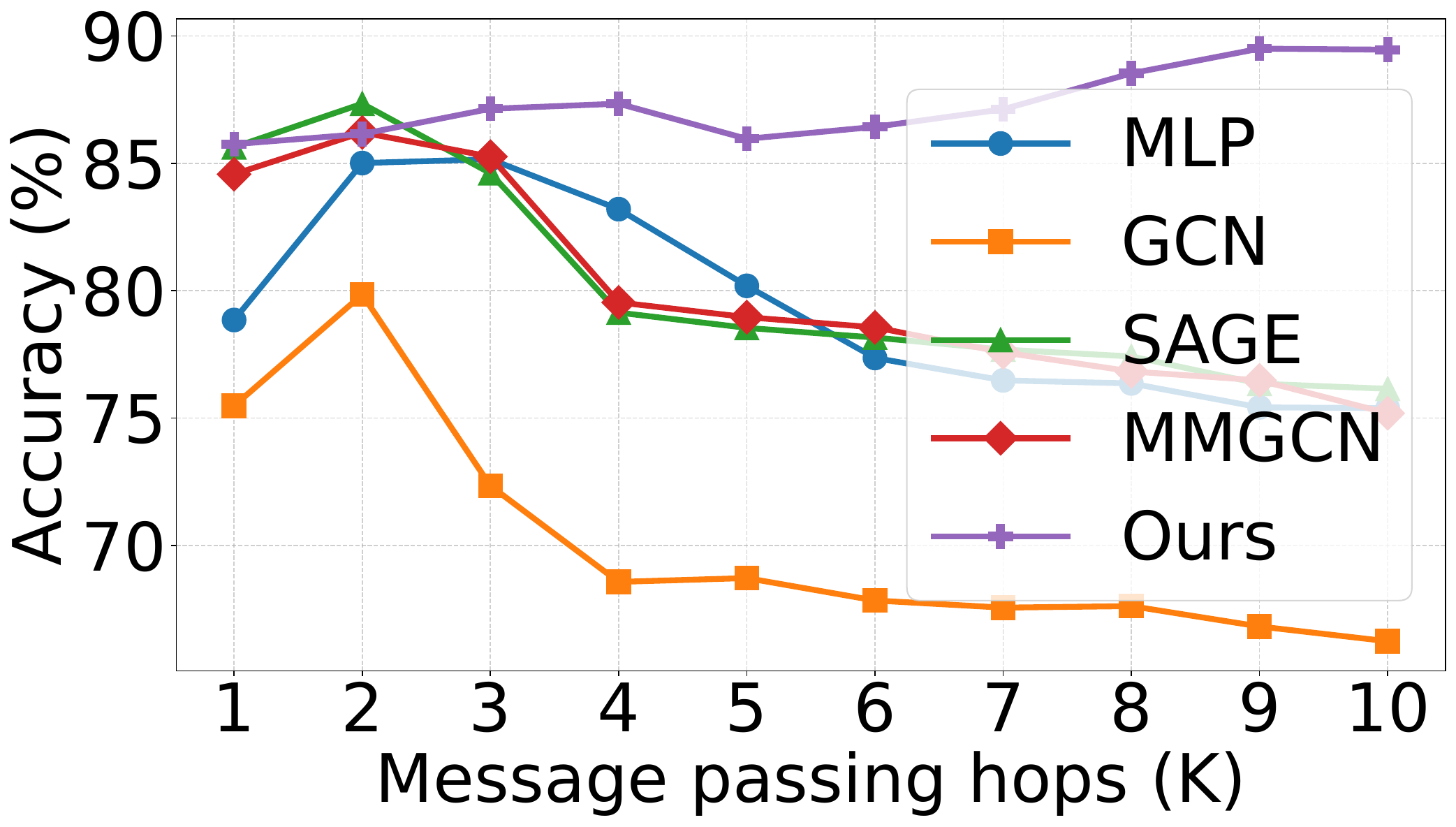}
    }
    \subfigure[The dynamic information pathways, path($\mathcal{V}^{(v)}, \mathcal{P}^{(v)}$) and ($\mathcal{V}^{(t)}, \mathcal{P}^{(t)}$)]{
        \label{fig:sim_pvt}
        \includegraphics[width=0.65\linewidth]{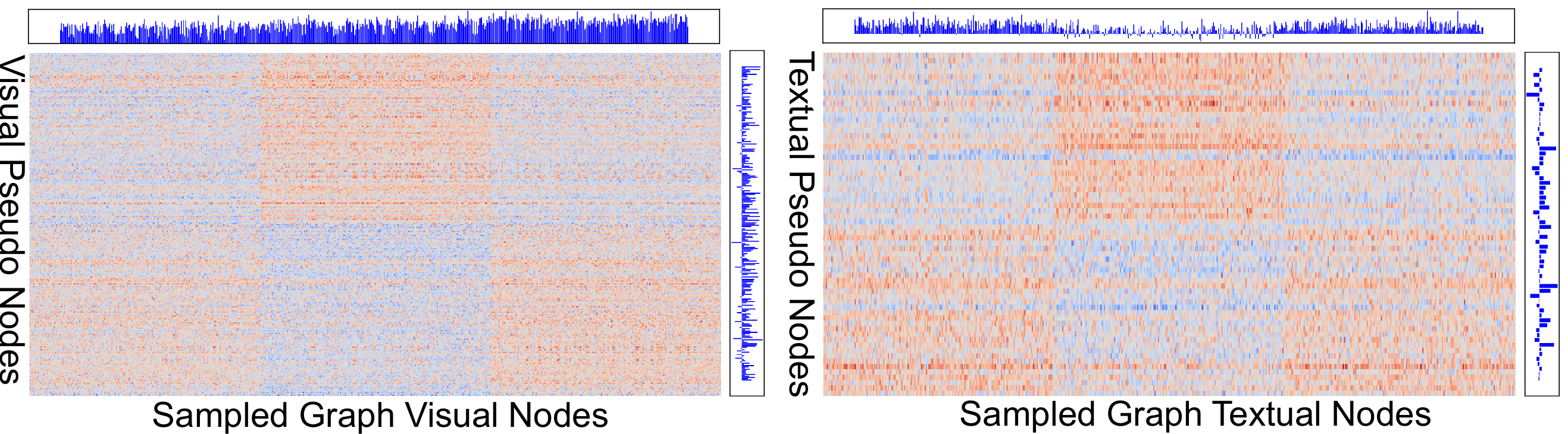}
    }
    \caption{(a) \mymodel with adaptive pathways can maintain the performance as the model depth increases. (b) Message passing pathways. The proximities between sampled graph nodes and pseudo nodes among two modalities. Some pseudo-nodes show link activation (bright rows), which may be a clustering pattern of nodes from different classes.}
\end{figure*}

The detailed link prediction results are presented in Table~\ref{tab:lp-results}, covering a diverse range of vision-language encoder combinations ($E_v, E_t$), including CLIP, ViT-T5, ImageBind-ImageBind, and DINOv2-T5. All metrics are reported as mean $\pm$ standard deviation over ten runs to ensure statistical reliability. As shown, our method consistently achieves state-of-the-art performance across all datasets and encoder configurations. In particular, it surpasses all baselines in MRR, Hit@1, and Hit@10 on Amazon-Sports, Amazon-Cloth, and Goodreads-LP, demonstrating both effectiveness and robustness. The most significant gains appear on Goodreads-LP, where our method exceeds the best baseline (e.g., BUDDY) by up to \textbf{+2.88} in MRR and \textbf{+5.79} in Hit@10, highlighting its strength in handling long-tail distributions and sparse relational structures. Comparable improvements are also observed under various encoder settings, suggesting that our design generalizes well to different visual and textual feature spaces. These results collectively confirm the effectiveness of our adaptive, modality-aware message passing mechanism in capturing intricate intra- and inter-modal dependencies, leading to more expressive and reliable multimodal graph representations.

\subsection{Multimodal Graph Node Classification}
\label{sec:exp_nc}

Table~\ref{tab:nc-results} reports node classification accuracy on Ele-Fashion and Goodreads-NC across various vision-language encoder settings. Our method consistently achieves the best performance under all configurations, outperforming both unimodal (e.g., MLP, GCN, SAGE) and multimodal baselines (e.g., MMGCN, MMGAT, UniGraph2). On Ele-Fashion, our model achieves up to \textbf{89.50\%} accuracy with ImageBind encoder, surpassing the strongest baseline by \textbf{+2.28\%}. On the more challenging Goodreads-NC, which features noisier and sparser labels, our method still leads across all encoder settings, achieving up to \textbf{85.37\%} in CLIP encoder and maintaining an average gain of \textbf{+2.35\%} over the best-performing baselines. These results demonstrate the generalizability and robustness of our approach to node classification tasks.

\begin{table}[t]
    \centering
    \resizebox{0.48\textwidth}{!}{
    \begin{tabular}{@{}l c cc c ccc@{}}
    \toprule
       \multirow{2}{*}{\scalebox{1.2}{Ablation}} && \multicolumn{2}{c}{Ele-Fashion} && \multicolumn{3}{c}{Amazon-Sports}\\
       
       \cmidrule(lr){3-4} \cmidrule(lr){6-8}
       && \textbf{Acc$\uparrow$} & \textbf{F1$\uparrow$} && \textbf{MRR$\uparrow$} & \textbf{Hit$@ 1\uparrow$} & \textbf{Hit$@ 10\uparrow$}\\
    \midrule
        w/o. $\mathcal{P}^{(v)}$ &&  84.62 & 78.48 && 31.24 & 17.26 & 69.85\\
        w/o. $\mathcal{P}^{(t)}$ && 85.34 & 77.68 && 31.59 & 17.42 & 70.16\\
        w/o. Local. && 84.62 & 77.81 && 30.26 & 16.85 & 61.34\\
        w/o. Global. && 83.16 & 76.45 && 29.48 & 16.37 & 67.48\\
        w/o. $\mathcal{P}^{(v)} \leftrightarrow \mathcal{P}^{(t)}$ && 82.47 & 74.16 && 30.17 & 16.03 & 65.27\\ 
        Ours && \textbf{87.88} & \textbf{82.06} && \textbf{34.26} & \textbf{18.45} & \textbf{72.54} \\
    \bottomrule
    \end{tabular}
    }
    \caption{The ablation studies of modules in \mymodel.}
    \label{tab:ablation}
\end{table}

\subsection{Ablation Study}
\label{sec:ablation_study}

We conduct ablation experiments to assess the contribution of each module in \mymodel. In Table~\ref{tab:ablation}, ``w/o $\mathcal{P}^{(v)}$'' and ``w/o $\mathcal{P}^{(t)}$'' remove visual and textual pseudo nodes; ``w/o Local'' and ``w/o Global'' disable local and global message passing; and ``$\mathcal{P}^{(v)} \leftrightarrow \mathcal{P}^{(t)}$'' removes cross-modal pseudo node interaction. From the ablation results, we have the following observations: (1) Message passing improves representation learning via structural context, confirming the benefit of relational modeling. (2) Pseudo nodes enhance intra-modal interaction. (3) Cross-modal communication complements embedding expressiveness. These findings confirm that each component of \mymodel is crucial for achieving strong and robust performance.

\begin{table}[t]
    \centering
    \resizebox{0.48\textwidth}{!}{
    \begin{tabular}{@{}l c cc c cc@{}}
    \toprule
       \multirow{2}{*}{\scalebox{1.0}{Model}} && \multicolumn{2}{c}{Ele-Fashion} && \multicolumn{2}{c}{Amazon-Sports}\\
       
       \cmidrule(lr){3-4} \cmidrule(lr){6-7}
       && Time(s) & Mem(MB) && Time(s) & Mem(MB) \\
    \midrule
       GCN && \textbf{0.253} & 1605.20 && \textbf{208.527} & 1962.37 \\
       SAGE && 0.326 & 1312.31 && 224.361 & 2043.06\\
       MMGCN && 1.237 & 2030.45 && 346.217 & 2570.52\\
       MGAT && 1.524 & 2340.30 && 382.425 & 2803.25\\
       Ours && 0.531 & \textbf{462.34} && 213.134 & \textbf{660.24}\\
    \bottomrule
    \end{tabular}
    }
    \caption{The model complexity analysis. Time (s) and memory (MB) consumed per epoch, the lower is better.}
    \label{tab:complex}
\end{table}

\subsection{Model Analysis}

\textbf{Tackling Over-Smoothing.} \mymodel alleviates over-smoothing by decoupling message passing from the input topology via adaptive pseudo nodes. Dynamic P2P connections enable diverse cross-modal aggregation, while the G2P subsystem preserves modality-specific signals across layers. This design supports deeper propagation without representation collapse. As shown in Figure~\ref{fig:over_smoothing}, \mymodel maintains higher Dirichlet energy than static baselines, verifying its ability to retain discriminative features and mitigate over-smoothing.

\begin{figure*}[t]
    \centering  
    \subfigure[($n_{p_v}, n_{p_t}$) on Ele-Fashion (Accuracy)]{
    \label{fig:npnodes_ele-fashion}
    \includegraphics[width=0.32\textwidth]{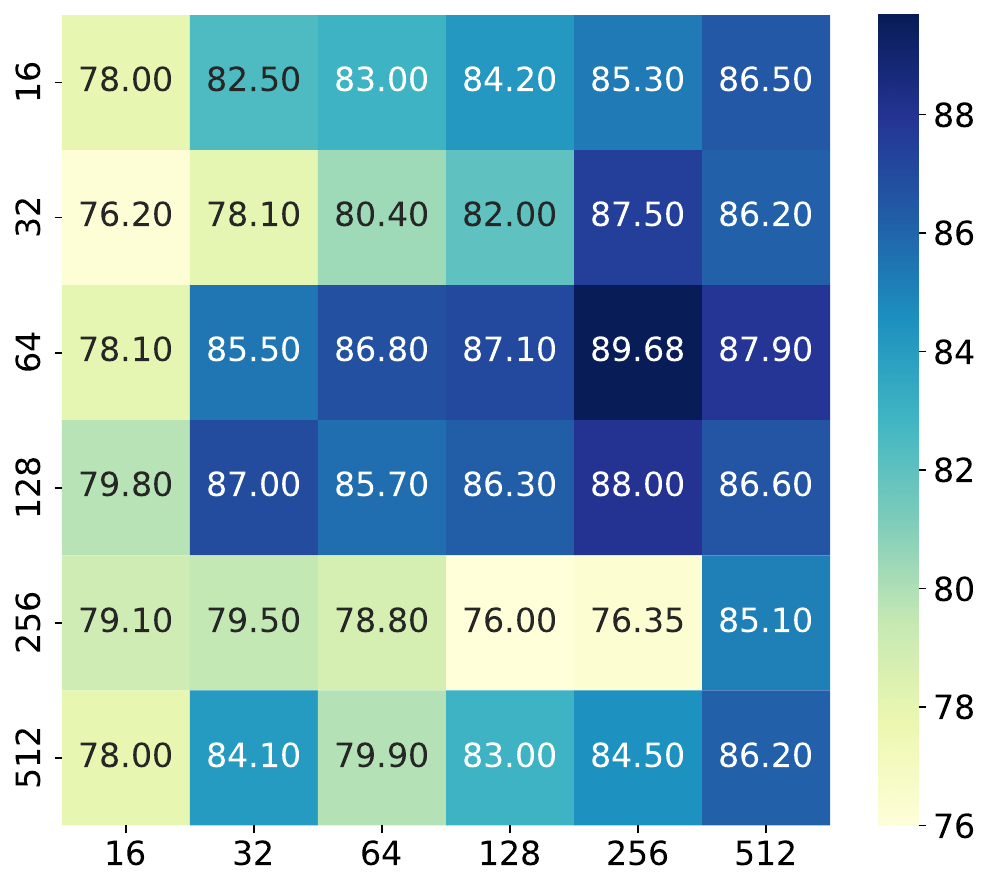}}
    \hfill
    \subfigure[($n_{p_v}, n_{p_t}$) on Amazon-Sports (MRR)] {
    \label{fig:npnodes_sports}
    \includegraphics[width=0.32\textwidth]{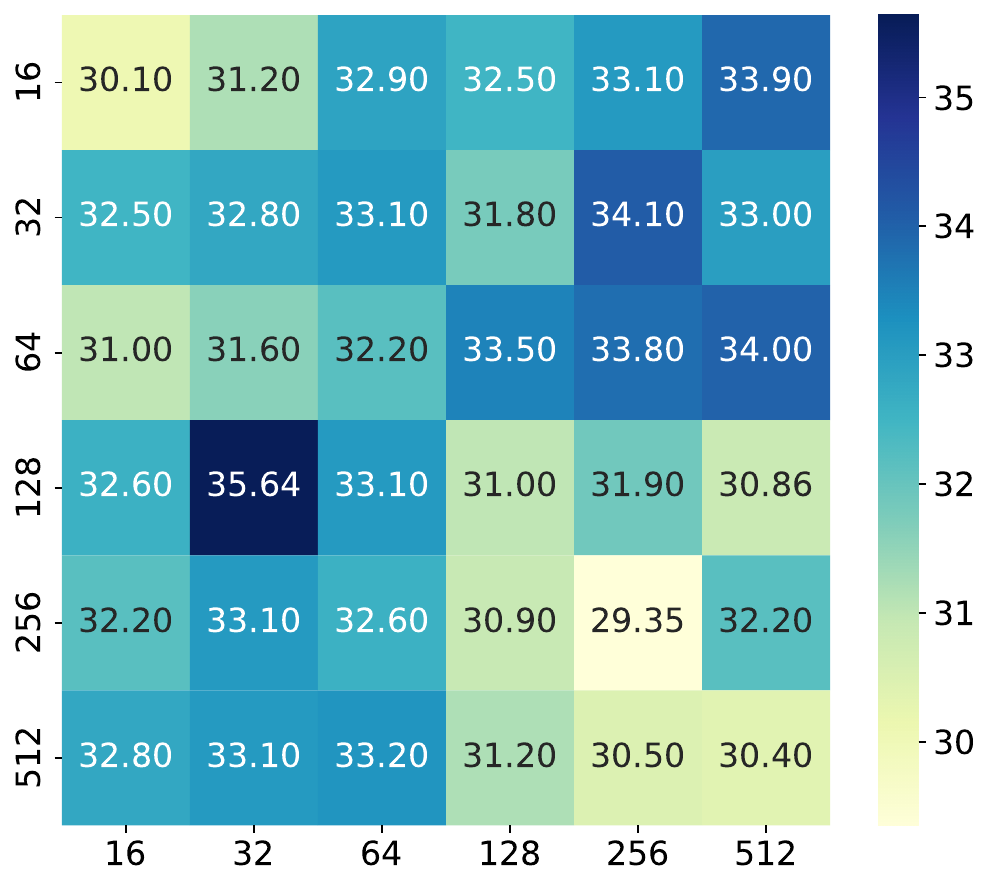}}
    \hfill
    \subfigure[($n_{p_v}, n_{p_t}$) on Amazon-Sports (Hit@1)] {
    \label{fig:npnodes_sports2}
    \includegraphics[width=0.32\textwidth]{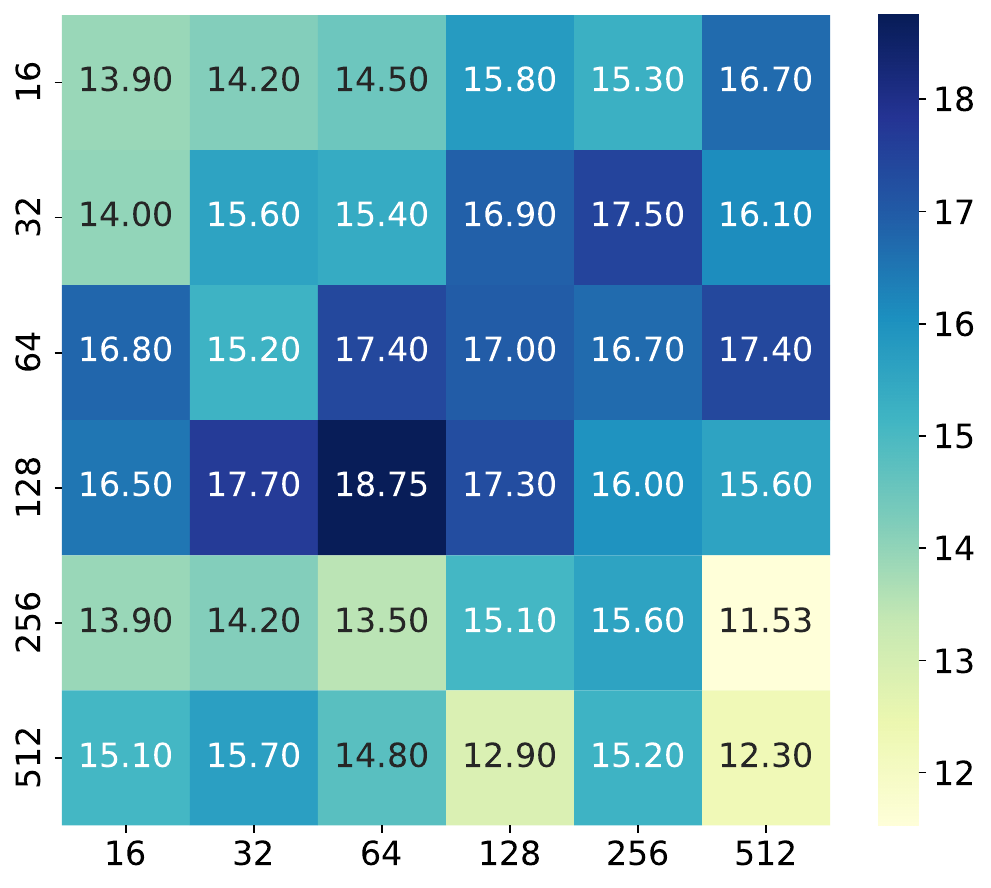}}
    \caption{Ablation on the number of visual ($n_{p_v}$) and textual ($n_{p_t}$) pseudo nodes.} 
    \label{fig:ablation_npnodes}
\end{figure*}

\begin{figure*}[t]
    \centering  
    \subfigure[SAGE]{
    \label{fig:sage_tsne}
    \includegraphics[width=0.23\textwidth]{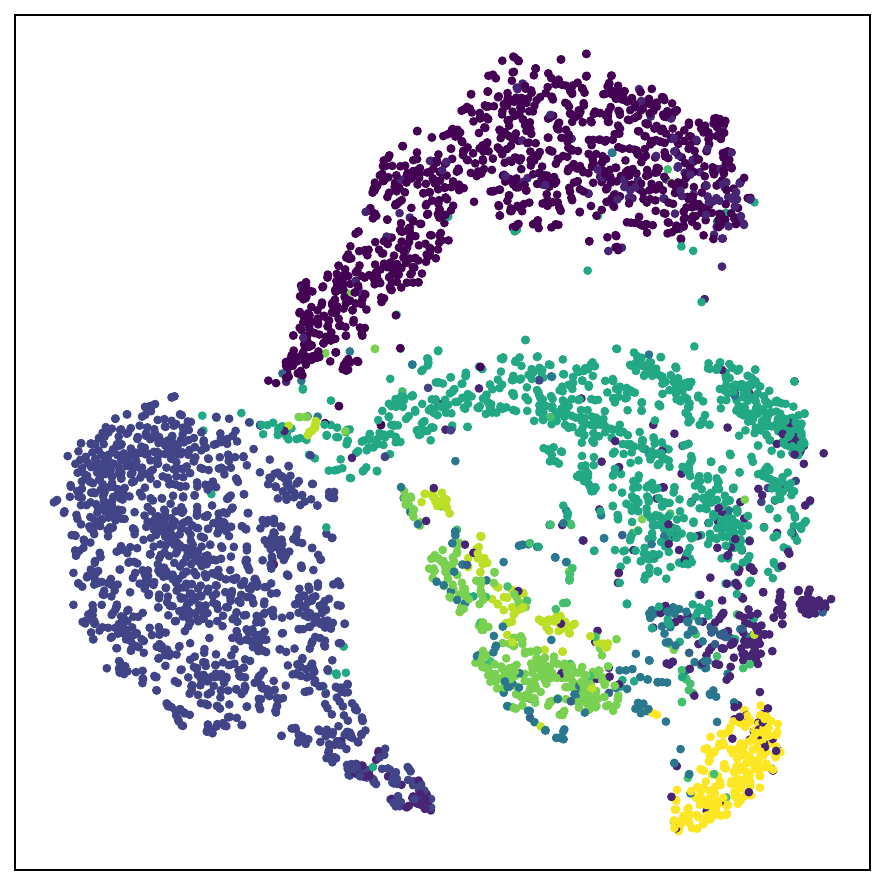}}
    \hfill
    \subfigure[MMGCN] {
    \label{fig:mmgcn_tsne}
    \includegraphics[width=0.23\textwidth]{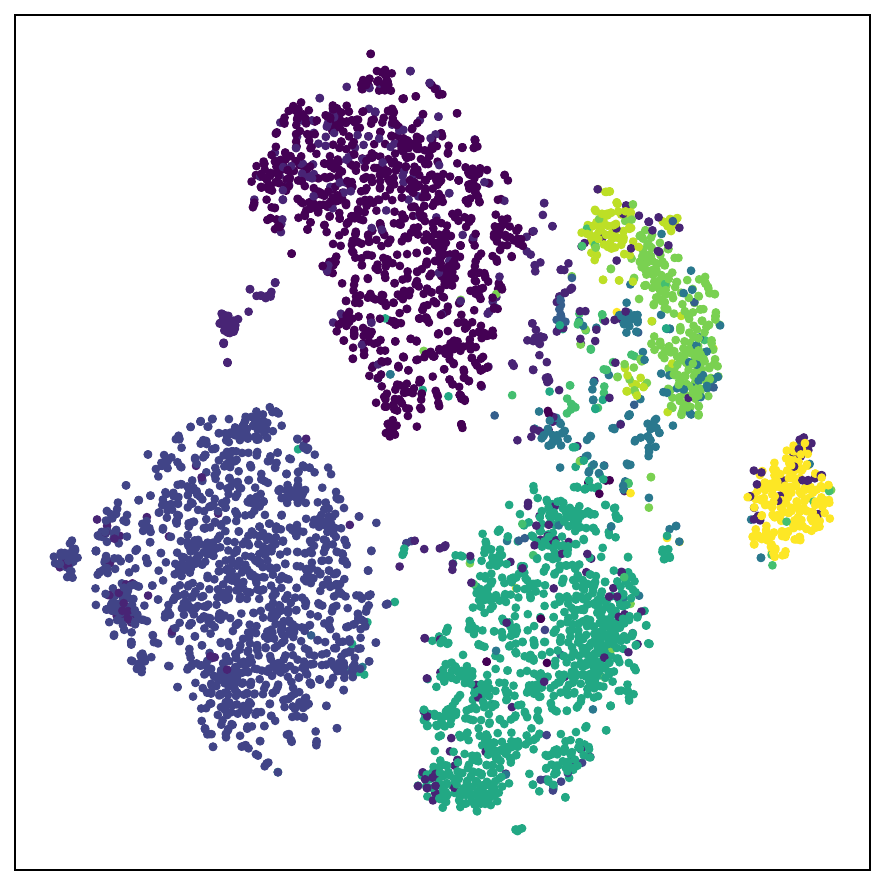}}
    \hfill
    \subfigure[MGAT] {
    \label{fig:mgat_tsne}
    \includegraphics[width=0.23\textwidth]{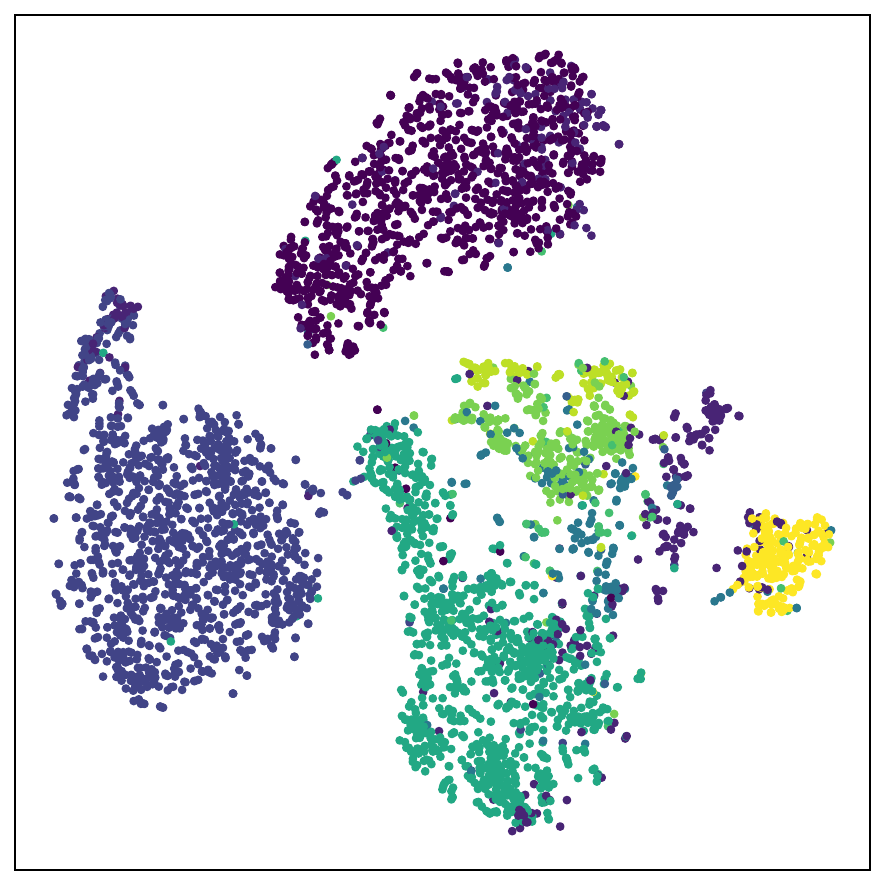}}
    \hfill
    \subfigure[\mymodel] {
    \label{fig:pammg_tsne_pnodes}
    \includegraphics[width=0.23\textwidth]{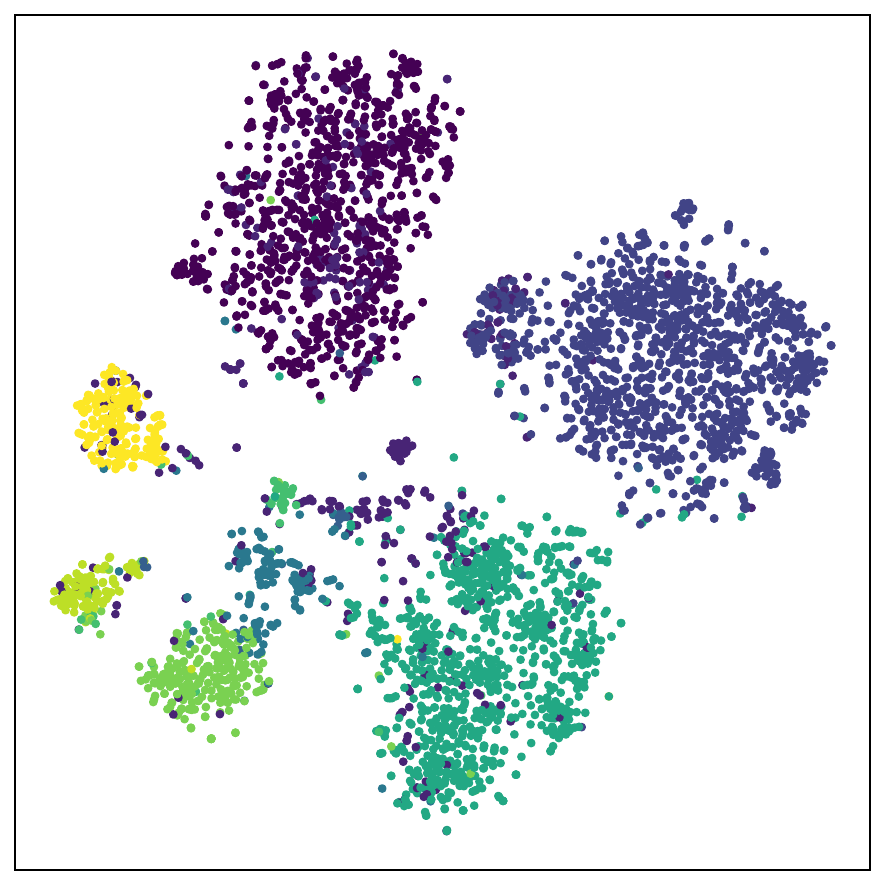}}
    \caption{The T-SNE plots for node embedding from Ele-Fashion dataset.} 
    \label{fig:tsne_fashion}
\end{figure*}

\textbf{Message Passing Pathways.} To understand the adaptive routing behavior of \mymodel, we visualize the similarity between sampled graph nodes and modality-specific pseudo nodes in Figure~\ref{fig:sim_pvt}. Notably, certain pseudo nodes exhibit strong activation patterns (bright rows), suggesting that they serve as hubs aggregating information from semantically related nodes across modalities. This clustering behavior indicates that \mymodel dynamically organizes message pathways based on latent structure rather than static topology.

\textbf{Complexity Analysis.} \mymodel performs adaptive message passing on multimodal graphs without dense pairwise modeling, with complexity $\mathcal{O}(\tau n n_p)$, where $n_p=\max(n_{p_v},n_{p_t})\ll n$, notably lower than uniform dense approaches. Empirical results in Table~\ref{tab:complex} show that our time complexity is comparable to efficient GNNs (e.g., GCN and SAGE), while memory overhead is significantly lower. These findings demonstrate that \mymodel achieves an effective balance between scalability and expressive power.

\textbf{Pseudo Node Numbers.} We perform a parameter search over the number of visual ($n_{p_v}$) and textual ($n_{p_t}$) pseudo nodes. As shown in Figure~\ref{fig:ablation_npnodes}, only a subset of pseudo nodes exhibit strong activation, while others remain underutilized. This indicates that using too many pseudo nodes may introduce redundancy, whereas too few may limit the model's expressiveness. We therefore select the optimal number of pseudo nodes based on validation performance to balance efficiency and representation capacity.

\subsection{Visualization}
\label{sec:visualization}
To qualitatively assess the expressiveness of node embeddings, we visualize the learned representations of different models using t-SNE. As shown in Figure~\ref{fig:tsne_fashion}, multimodal GNNs produce more structured and discriminative embeddings compared to their unimodal counterparts, particularly for nodes associated with multiple modalities. Notably, our proposed method yields superior category separability, with clearer decision boundaries and reduced embedding overlap. This suggests that \mymodel better captures modality-specific semantics while promoting cross-modal alignment, which is especially beneficial for distinguishing classes with ambiguous or fuzzy boundaries. These visualization results align with our quantitative gains and provide further evidence for the effectiveness of our multimodal message passing design.

\section{Conclusion}
\label{sec:conclusion}

 In this work, we propose \mymodel, a novel pseudo node-enabled framework for multimodal graph representation learning with dynamic information pathways. By decoupling message propagation from fixed graph topology and leveraging modality-aware pseudo nodes, \mymodel effectively captures both intra- and inter-modal dependencies while maintaining scalability. Extensive experiments on link prediction and node classification tasks demonstrate its superiority over strong baselines in both performance and efficiency. Our analysis further shows that \mymodel mitigates over-smoothing and adapts message pathways dynamically, offering a promising direction for future research on structured multimodal learning.

\section{Acknowledgments}
This work was supported by Basic Research Program of Jiangsu (Grant No. BK20251198), the Natural Science Foundation of Jiangsu Province (Grant No. BK20222003), the National Natural Science Foundation of China (Grant Nos. 62502201, 62572236), Postdoctoral Fellowship Program of CPSF (Grant No. GZC20251067), Jiangsu Funding Program for Excellent Postdoctoral Talent, the Collaborative Innovation Center of Novel Software Technology and Industrialization, and the Sino-German Institutes of Social Computing.

\bibliography{reference}

\begin{figure*}[th]
    \centering
    \includegraphics[width=\linewidth]{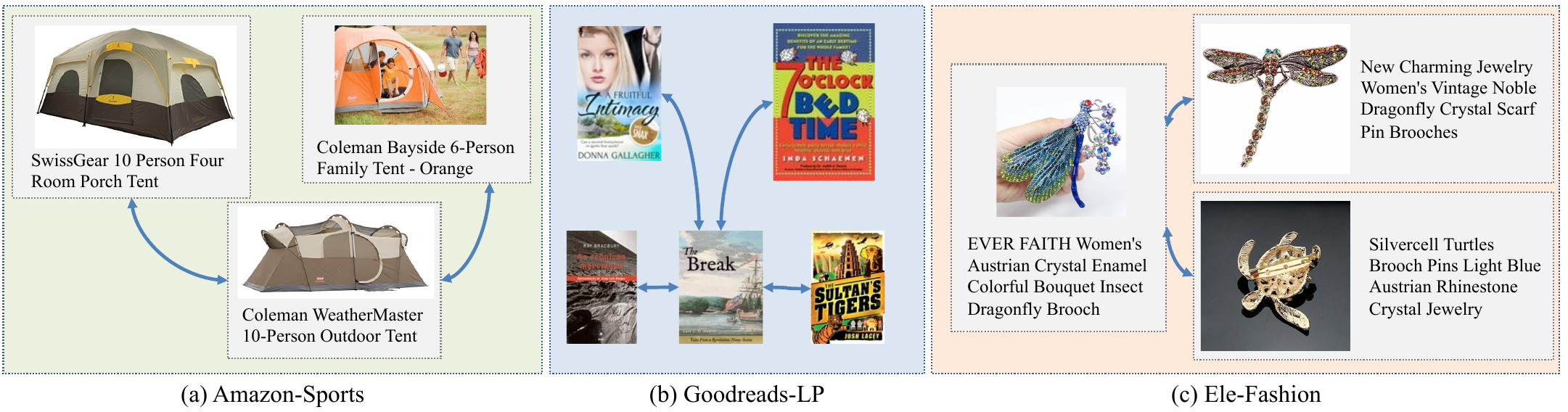}
    \caption{Visualization of multimodal graph benchmarks. All nodes of these benchmarks have both visual and text features. \textbf{(a) Amazon-Sports:} The image and text come from the original image and title of the sports equipment. \textbf{(b) Goodreads-LP:} The images correspond to book covers. We do not show the text features of Goodreads-LP since the book description is very long. \textbf{(c) Ele-fashion:} The images and texts correspond to the original image and title of the fashion product, respectively.}
    \label{fig:teaser}
\end{figure*}

\section{Dataset Information}
\label{sec:dataset}

\textbf{Amazon-Sports} and \textbf{Amazon-Cloth} are two benchmark datasets for link prediction, constructed from the larger Amazon-Review corpus~\cite{ni2019justifying, hou2024bridging}. In these datasets, each node represents a distinct product listed on Amazon within either the sports or clothing category, while edges denote co-purchasing behavior—indicating that two items were frequently bought together by users. Each product is associated with multimodal information: textual features derived from its product title and visual features sourced from its high-resolution image. As illustrated in \cref{fig:teaser}(a), Amazon-Sports contains nodes such as tents with varied shapes and appearances, captured against diverse natural backgrounds, while their corresponding textual descriptions provide additional semantic cues. To prepare the datasets, we extracted the relevant metadata for each item along with their co-purchase links, ensuring both structural and modal information are preserved for downstream learning tasks.

\textbf{Goodreads-LP} and \textbf{Goodreads-NC} are derived from the Goodreads Book Graph dataset~\cite{wan2018item, wan2019fine}, and are designed for link prediction and node classification tasks, respectively. In these graphs, each node represents an individual book listed on Goodreads, while edges reflect user preference patterns—specifically, whether readers who liked one book are likely to enjoy another, as established in~\cite{zhu2024touchup}. Each book is associated with multimodal attributes: the textual modality consists of the book’s description, offering insight into its content and genre, while the visual modality corresponds to its cover image. To ensure the availability of both modalities, we filter out books lacking cover images during preprocessing. This results in a multimodal graph that preserves user-driven relational structures alongside rich semantic and visual information for each item.

\textbf{Ele-Fashion} is a node classification dataset constructed from the Amazon-Fashion collection~\cite{ni2019justifying, hou2024bridging}. In this dataset, each node represents a fashion-related product available on Amazon, and edges indicate co-purchasing behavior—signifying that the connected items are frequently bought together. The dataset incorporates multimodal node features: textual attributes are derived from the product titles, capturing key descriptive elements such as category and style, while visual features consist of the corresponding product images. This multimodal setup enables models to jointly leverage semantic and visual cues to predict product categories.
We summarize the dataset abstract in Table~\ref{tab:datasets-overview}, including the text/visual features, task, metrics, scale, and split ratio. The detailed graph-based statistics of datasets are reported in Table~\ref{tab:datasets-stats}.

\begin{table*}[t]
\caption{Overview of the multimodal graph datasets. We present five multimodal graph datasets with varying scale and tasks. All datasets are split at random. Abbreviations: \underline{LP} = Link Prediction. \underline{NC} = Node Classification. \underline{OR} = Original.}
\label{tab:datasets-overview}
\centering
\setlength{\tabcolsep}{10pt} 
{\small
\begin{tabular}{lrrrrrr}
\toprule

\textbf{Name} & \textbf{Text Features} & \textbf{Visual Features} & \makecell{ \textbf{Task}} & \textbf{Metrics} & \textbf{Scale} & \textbf{Split Ratio}\\
\midrule

Amazon-Sports & Product Titles & Product Images & LP & MRR, Hits@K & Small & 8/1/1 \\

Amazon-Cloth & Product Titles & Product Images & LP & MRR, Hits@K & Medium & 8/1/1 \\

Goodreads-LP & Book Description & Book Images & LP & MRR, Hits@K & Large & 6/1/3\\

Ele-Fashion & Fashion Titles& Fashion Images & NC & Accuracy & Medium & 6/1/3
\\

Goodreads-NC & Book Description & Book Images & NC & Accuracy & Large & 6/1/3
\\

\bottomrule
\end{tabular}
}
\end{table*}

\begin{table*}[t!]
\caption{Detailed graph-based statistics of datasets. 
\underline{CC} = Cluster Coefficient. \underline{RA} = Resource Allocation. \underline{N/A} = Not Applicable (Nodes do not have class labels).}
\label{tab:datasets-stats}
\centering
\resizebox{0.88\textwidth}{!}
{
\begin{tabular}{l@{\hskip5pt}r@{\hskip5pt}r@{\hskip5pt}r@{\hskip5pt}r@{\hskip5pt}r@{\hskip5pt}r@{\hskip5pt}r}
\toprule
   \textbf{Name} &  \textbf{Nodes} & \textbf{Edges}  & \textbf{Average Degree} & \textbf{Average CC}  & \textbf{Average RA} & \textbf{Transitivity} & \textbf{Edge Homophily}\\
\midrule 

     Amazon-Sports & 50,250 &  356,202 & 14.18  &  0.4002 &  0.3377 & 0.2658 & N/A\\ 

    Amazon-Cloth &  125,839 & 951,271 &15.12 & 0.2940 &  0.2588  & 0.1846 & N/A\\

   Goodreads-LP & 636,502 & 3,437,017 & 10.79 & 0.1102 & 0.0685 & 0.0348 & N/A \\ 
   
    Goodreads-NC & 685,294 & 7,235,084 & 21.11 & 0.1614 &  0.1056 & 0.0498 & 0.6667
    \\

    Ele-Fashion & 97,766 & 199,602 & 4.08 & 0.1730 & 0.1467& 0.0560 & 0.7675
    \\

\bottomrule
\end{tabular}

\vspace{-2em}
}
\end{table*}

\textbf{Data Availability and Ethics.}
These benchmarks are organized from existing open source data, with proper open source licenses. 
Amazon-Sports and Amazon-Cloth and Ele-Fashion are available with Apache License~\footnote{\url{https://github.com/PeterGriffinJin/Patton}}.
Goodreads-LP, Goodreads-NC are released under MIT License~\footnote{\url{https://mengtingwan.github.io/data/goodreads.html}, \url{https://github.com/tsafavi/codex/tree/master}}.
These do not involve interaction with humans or private data.

\section{Baselines}
\label{sec:baseline}

To enable a comprehensive evaluation of GNN architectures on multimodal graph data, we include five representative GNN models: GCN~\cite{kipf2016semi}, SAGE~\cite{hamilton2017inductive}, MMGCN~\cite{wei2019mmgcn}, MGAT~\cite{tao2020mgat}, and BUDDY~\cite{chamberlain2022graph}. Additionally, following ~\cite{hu2020open}, we report the performance of an MLP as a baseline to evaluate the usefulness of graph structure.

\textbf{Conventional GNNs.} We include several representative baseline models—GCN, SAGE, and MLP—for both link prediction and node classification tasks. \textbf{GCN}\cite{kipf2016semi} performs message passing through neighborhood aggregation based on spectral graph convolutions. \textbf{SAGE}\cite{hamilton2017inductive} generalizes this idea by sampling and aggregating information from a node’s local neighborhood using learnable functions, supporting inductive learning. \textbf{MLP}, a non-graph baseline, processes node features independently without exploiting the graph structure. For all unimodal baselines, we concatenate the text and image embeddings to form a joint input before feeding them into the respective models. We exclude \textbf{BUDDY}~\cite{chamberlain2022graph} from node classification experiments, as it is tailored specifically for link prediction scenarios and is not applicable in node-level classification settings.

\textbf{Multimodal GNNs.}
Although traditional GNNs have demonstrated strong capabilities on graphs with unimodal features, adapting them to effectively process multimodal graph data remains an underexplored area. To benchmark the performance of GNNs in the multimodal setting, we incorporate and adapt two representative models originally proposed for recommendation tasks:

(1) \textbf{MMGCN}~\cite{wei2019mmgcn} builds separate user-item graphs for each modality (e.g., text and image), and applies GNNs independently to each to capture intra-modality interactions. The user and item embeddings from different modalities are subsequently fused at the final prediction stage. For our experiments, we adapt MMGCN to link prediction by employing a dot product decoder over the fused embeddings. For node classification, we append a 1-layer MLP to project the multimodal embeddings into the label space.

(2) \textbf{MGAT}~\cite{tao2020mgat} extends the Graph Attention Network (GAT) framework by introducing modality-specific attention heads. Each modality is first encoded through a separate attention mechanism, and the resulting representations are aggregated via a cross-modal attention layer that dynamically weighs the importance of each modality. As with MMGCN, we use a dot product decoder for link prediction and a 1-layer MLP for node classification.

(3) \textbf{UniGraph2}~\cite{he2025unigraph2} is a foundation model designed for representation learning on multimodal graphs. It integrates modality-specific encoders with a shared GNN to jointly model multimodal node features and structural relationships. To support generalization across different domains, UniGraph2 introduces a cross-domain multi-graph pretraining strategy and incorporates a Mixture of Experts (MoE) module to align features from heterogeneous modalities. For consistency with other models, we use dot product for link prediction and a 1-layer MLP for node classification.

\section{Modality Encoders}
\label{sec:encoders}

\textbf{Text Encoders.} We consider three representative text encoders: \textbf{T5}\cite{raffel2020exploring}, \textbf{CLIP}\cite{radford2021learning}, and \textbf{ImageBind}~\cite{girdhar2023imagebind}. T5 is a transformer-based encoder-decoder model pre-trained on a large-scale text-to-text objective, and is widely used for generating high-quality text embeddings. CLIP jointly trains text and image encoders to align their embeddings in a shared space using contrastive learning, making it suitable for vision-language tasks. ImageBind further generalizes this idea by learning a shared embedding space across six modalities, including text and image, enabling broader cross-modal alignment.

\textbf{Visual Encoders.} We include four visual encoders in our study: \textbf{ViT}\cite{dosovitskiy2020image}, \textbf{DINOv2}\cite{oquab2023dinov2}, \textbf{CLIP}\cite{radford2021learning}, and \textbf{ImageBind}\cite{girdhar2023imagebind}. ViT is a transformer-based model trained with full supervision on large-scale image classification tasks. DINOv2 adopts a self-supervised learning framework to learn high-quality visual features without labeled data. CLIP learns visual representations aligned with language by jointly training image and text encoders in a contrastive manner. ImageBind extends this alignment to additional modalities by mapping multiple sensory inputs—including images, text, and audio—into a shared embedding space, offering broader potential for multimodal graph applications.

Our choice of feature encoders allows us to explore key aspects of multimodal graph learning, including:

\begin{itemize}
\item The role of modality alignment in creating a unified embedding space across different data types.
\item A comparison between supervised and self-supervised approaches for learning feature representations.
\end{itemize}

\end{document}